# Object Detection in Thermal Spectrum for Advanced Driver-Assistance Systems (ADAS)

Muhammad Ali Farooq, Peter Corcoran, Fellow, IEEE, and Cosmin Rotariu, Waseem Shariff

*Abstract*—**Object detection in thermal infrared spectrum provides more reliable data source in low-lighting conditions and different weather conditions, as it is useful both in-cabin and outside for pedestrian, animal, and vehicular detection as well as for detecting street-signs & lighting poles. This paper is about exploring and adapting state-of-the-art object detection and classifier framework on thermal vision with seven distinct classes for advanced driver-assistance systems (ADAS). The trained network variants on public datasets are validated on test data with three different test approaches which include test-time with no augmentation, test-time augmentation, and test-time with model ensembling. Additionally, the efficacy of trained networks is tested on locally gathered novel test-data captured with an uncooled LWIR prototype thermal camera in challenging weather and environmental scenarios. The performance analysis of trained models is investigated by computing precision, recall, and mean average precision scores (mAP). Furthermore, the trained model architecture is optimized using TensorRT inference accelerator and deployed on resource-constrained edge hardware Nvidia Jetson Nano to explicitly reduce the inference on GPU as well as edge devices for further real-time onboard installations.**

## I. Introduction

ADVANCE Driver-Assistance Systems (ADAS) has become an emerging consumer technology application and the evolution of this technology over time aims to provide extended safety benefits and reliable means of transportation. Various key technologies are directly associated with ADAS which includes, sensor fusion for real-time data logging, and object/ obstacle detection and tracking system using machine learning algorithms. This will empower the drivers to monitor the external environment, detecting external objects, and predict events that the driver needs to be aware of thus providing a deeper understanding of the entire road network.

Current ADAS largely rely on computer vision and machine learning which uses visible (RGB) or RGB + near-infrared (NIR) cameras as a sensor; the alternative is lidar and radar based [1]. Practical systems often leverage both camera + radar and lidar. However, the mentioned sensors and imaging modalities have some of their limitations. For instance, lidar provides a sparse three-dimensional (3D) map of the environment, but small objects like pedestrians and cyclists are difficult to detect especially when they are at a distance [2]. The RGB camera operates inadequately in unfavorable illumination conditions such as low lighting, sun glare, and glare from the headlight beam. Radar has a low spatial resolution to detect pedestrians accurately [2]. Also, large objects such as cars can saturate the performance of the receiver if they are closer to the transmitter, and lastly, the performance of the radar is severely affected as the radio signals can face enough natural interference.

Recent developments in bolometer technology have led to lower costs for thermal imaging sensors. These sensors can complement or even replace existing technology, offering a particular advantage that as they sense thermal emissivity of objects, they operate independently of lighting conditions. They can also offer in-cabin solutions for driver and occupant monitoring and could be employed for enhanced street-scene understanding thus making it a more consistent solution for smart-city environments.

Thermal imaging is a non-contact method in which the thermal radiations of an object is converted into a visible image also referred to as a thermogram. Thermal cameras can operates in different wavelength spectrums such as long-wave (LWIR) and middle-wave (MWIR) and can be used for object detection in both day and night-time environmental conditions [46]. Since it is invariant to illumination changes, occuluisons and shadows it provides improved situational awareness that results in more robust, reliable, and safe ADAS. Moreover, by integrating with state-ot-art AI-based imaging pipelines we can detect multiple objects of different classes for advanced driver assistance & monitoring systems. In this study, we have primarily focused on thermal object detection using state-of-the-art YOLO-V5 [3, 4] end-to-end deep learning framework as it can play a key

October 9, 2021, "This project (Link: https://www.heliaus.eu/) has received funding from the ECSEL Joint Undertaking (JU) under grant agreement No 826131. The JU receives support from the European Union's Horizon 2020 research and innovation program and National funding from France, Germany, Ireland (Enterprise Ireland International Research Fund), and Italy". The authors would like to acknowledge Cosmin Rotariu from Xperi-Ireland and rest of team members for providing the support in preparing the data accusation setup and helping throughout in data collection and Quentin Noir from Lynred France for giving their feedback. Moreover, authors would like to acknowledge the contributors of the all the public datasets for providing the image resources to carry out this research work and ultralytics for sharing the YOLO-V5 Pytorch version.

Muhammad Ali Farooq and Waseem Shariff is with the National University of Ireland Galway (NUIG), Galway, H91TK33, Ireland (e-mail: m.farooq3@nuigalway.ie). Peter Corcoran is with Department of Engineering, National University of Ireland Galway (NUIG), Galway, H91TK33, Ireland (e-mail: peter.corcoran@nuigalway.ie).

Cosmin Rotariu is working with XPERI Corporation, H91VY1, Ireland (email: cosmin.rotariu@xperi.com).

Project video link: https://cutt.ly/Ybt6ZBw

element in the successful implementation and deployment of ADAS. The main goal is to achieve robust training results using two different optimizers (SGD and ADAM) and rigorous network testing using three different approaches which includes test-time with no augmentation (TTNA), test-time augmentation (TTA), and test-time with model ensembling (TTME) are incorporated for improved test accuracy.

*A. Main Contributions of This Paper*

The core contributions of this research work includes

- Adaptation and validation of state-of-the-art object detection/ classification framework on thermal imaging for ADAS with seven distinct classes which includes stationary as well as moving objects.
- A novel test dataset is captured in two different methods using a uncooled LWIR thermal camera developed under Heliaus EU project [37] in different environments and weather conditions. A total of 20,000 thermal frames have been selected for this study consisting of different class objects.
- Evaluating a neural framework with a range of model sizes to determine its suitability for porting to a resource-constrained embedded edge platform (Nvidia Jetson). Thus, to study its feasibility for further automotive on-board-computer (OBC) installations [42].

## II. BACKGROUND/ RELATED WORK

Common practices in ADAS architecture have been established over the years. Most of these systems divide the task of safe and advanced driving into subcategories and employ an array of sensors and algorithms on various hardware modules for diversified tasks. More recently, end-to-end driving started to emerge as an alternative to modular approaches. Conventional machine learning and deep learning models [36] have become dominant in many of these tasks among which object detection is one of them. This section will mainly focus on the published studies exploring state-of-the-art object detection methods, and the reported results that investigate the area of object detection in the thermal spectrum. It includes various object classes such as pedestrian and vehicle detection.

*A. Object Detection using Conventional Machine Learning Algorithms*

Conventional machine learning methods mainly rely on manually extracted feature vectors which are then fed different types of classifiers and detector for performing object detection either offline or in real-time. Olmeda et al. [18], proposes pedestrian detection in FIR images by using a new feature descriptor, the histograms of oriented phase energy (HOPE), and an adaptation of the latent variable based on the support vector machine (SVM). HOPE is a contrast invariant descriptor that encodes a grid of local-oriented histograms extracted from the phase congruency of the images computed from a joint of Gabor filters. The authors concluded that histogram-based features perform exceptionally well as compared to other Linear binary patterns (LBP) and Principle component Analysis (PCA). Similarly, in [19] authors have focused on detecting pedestrians using night-time thermal imaging. The overall system is based on image segmentation, ROI selection, and histograms of oriented gradients (HOG) [20] for feature extraction. In the next step Adaboost classifier [21] is used for pedestrian detection. According to the authors the same method can be used for the development of the intelligent driver assistance system, giving more road traffic situations to the drivers throughout the night-time. In another study by Soundrapandiyan et al. [22] authors had employed simple image processing methods which include background subtraction and adaptive thresholding is applied for person detection in the thermal spectrum. The authors achieve an overall accuracy of 90% in different environmental conditions. Besbes et al. [45] propose a pipeline for pedestrian detection in thermal images by using a hierarchical codebook of Speeded Up Robust Features (SURF) in the head region, taking advantage of the brightness of this area inside the regions of interest (ROIs). The reported experimental results show improved accuracy as compared to Haar-like Adaboost-cascade, linear SVM, and MultiFtr pedestrian detectors, trained on the FIR images. Coming towards more recent studies Lahmayed et al. [23] presented a method based on three different feature extractors. It includes multi-threshold and Histogram of Oriented Gradients (HOG) and Histograms of Oriented Optical Flow (HOOF) colour features combined with an SVM using both thermal infrared and visible light images. The authors validated their algorithm on three different datasets i.e., OSU colour thermal dataset [14], video analytic dataset, and LITIV dataset [24].

*B. Object Detection in Thermal Spectrum using Deep Learning*

Deep learning and most specifically convolution neural networks (CNN) have become an emerging trend for building artificially intelligent imaging pipelines due to its robustness and precision accuracy. It has proved its strengths and gained more interest in object detection using visible spectrum data. There are various state-of-the-art published deep learning based object detection frameworks which include YOLO [5], Single Shot MultiBox Defender (SSD) [7], R-CNN [8], Fast R-CNN [9], and Mask R-CNN [10]. All of these networks are build using an end-to-end deep learning network. The efficacy of these algorithms is mostly tested RGB datasets which include MS COCO dataset [11], ImageNet [12], and PASCAL-VOC [13]. However, in this study, we are more interested to explore the robustness of the aforementioned object detector i.e. YOLO in thermal infrared spectrums. We can find various published studies [25-29] where deep learning is employed for object detection in thermal images. In these studies, authors have used thermal data for object detection i.e., pedestrian detection in differing illumination conditions. The system works by extracting the feature maps from multispectral images. In the next step, these feature maps are fed to state-of-art object detectors which include faster-RCNN [30] and YOLO [5]. Herrmann et al. [31] tested the Single Shot Detector (SSD) object detector by applying different pre-processing techniques to assess the performance of the detector on thermal data. They used KAIST [15] dataset for performance evaluation. The authors also worked with Maximally Stable Extremal Regions (MSERs) and later on classify the detected proposals by using CNN. The approach was tested on the OSU thermal pedestrian [14], OSU colour thermal [14], and Terravic motion IR [32] datasets. Recently, Huda, Noor Ul, et al. [33] used a YOLO object detector for person detection in the thermal spectrum. The authors had created their outdoor thermal dataset for transfer learning the YOLO-v3. The trained models were tested on three different public datasets which include CVC-09 [17], OSU-Thermal [14], and BU-TIV-atrium [34]. Similarly, in another recent study by Munir, Farzeen, et al. [35] authors have explored domain adaption through style transfer methodology. Authors have used GAN architectures and cross-domain models on thermal and visible spectrum images. In the next stage, the style consistency approach is used for object detection by using two different public datasets, FLIR ADAS [16] and Kaist Multi-spectral dataset [15]. It is difficult to find published studies where authors have investigated the real-time feasibility of object detection algorithms in the thermal spectrum on edge devices for ADAS applications.

### III. PROPOSED METHODOLOGY

This section mainly focuses on the proposed methodology for robust training of state-of-the-art Yolo-v5 framework [3, 4] for out-cabin object detection in the thermal spectrum. Yolo was first introduced by Redmon, Joseph, et al. [5] for real-time object detection. YOLO is considered as one of the fastest and finest deep learning algorithms for object detection in images, videos, and real-time camera streams. It can process up to 45 frames per second. The algorithm utilizes regression techniques thus training the whole image at once to optimize the overall performance. Moreover, it detects the class objects with their probabilities scores at the same time without requiring region proposals. YOLO-V5 is natively implanted in PyTorch whereas all prior models in the YOLO family leverage Darknet deep learning framework [6]. The networks are trained to detect seven different objects which include pedestrian/ person, vehicles (car, bus, bike, and bicycle), animal (dog), and light/ sign poles. All these objects are commonly found on the roadside thus it will provide a better perspective for the driver's assistance. In this study, we have reviewed four large-scale datasets in the thermal domain. These datasets are available publicly and provide image sources with differing outdoor environmental and weather conditions. These datasets include the OSU Thermal pedestrian [14] database, KAIST Multi-Spectral dataset [15], FLIR ADAS dataset [16], and CVC-09 [17] datasets. Fig. 1 shows the sample images from these datasets under different environmental and lighting conditions.

Table I provides the complete dataset attributes of all four datasets used in this study. The selected set of public datasets is used for optimal training and testing of four different network variants of YOLO-V5 named as X-large, large, medium, and small models. Most of these datasets are specifically gathered and proposed for autonomous driving applications. Fig. 2 shows the complete block diagram representation of the proposed methodology used in this study. We have used data samples from four different public datasets as shown in Table 1 for the training of all four network variants of YOLO-V5 architecture. The numeric performance comparison of all the model variants has been evaluated in the experimental results and discussion section thus summarizing the best models in terms of precise accuracy and lower inference time. The trained networks are tested on both public as well as locally gathered datasets.

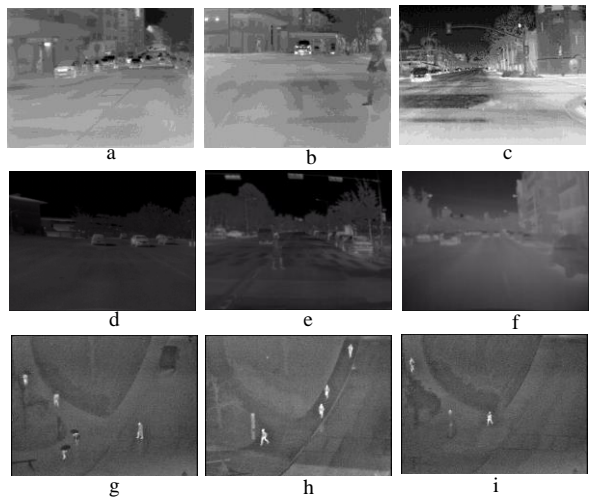

Fig. 1. Sample images under varying environmental conditions from four different public datasets (a) CVC-09 dataset: frame acquired in day time road environment, (b) CVC-09: frame acquired in night-time road environment, (c) FLIR ADAS dataset: frame acquired in cloudy weather and road environment, (d) KAIST dataset: frame acquired in day time campus environment, (e) KAIST dataset: frame acquired in day time road environment, (f) KAIST dataset: frame acquired in night-time road environment, (g) OSU-thermal dataset: frame acquired in day-time with light rain in Ohio State University campus, (h) OSU-thermal dataset: frame acquired in day-time with partly cloudy weather in the campus environment and (i) OSU-thermal dataset: frame acquired in day-time with haze/ dusty weather conditions in the university campus environment.

TABLE I
Datasets Attributes

| Datasets | Weather Conditions and viewpoint | | Environment | No of Frames | Objects | Camera specifications and image resolution |
|---|---|---|---|---|---|---|
| CVC-09 [17] | Daytime and nighttime nearside | | roadside | 11,000 frames | Person Cars Poles Bicycle Bus Bikes | 640 x 480 |
| FLIR ADAS [16] | Daytime and nighttime with Sun and cloudy condition Nearside | | roadside | 14,000 frames | Person Cars Poles Bus Dog Bicycles | FLIR Tau2 LWIR camera 640 x 512 |
| OSU Thermal [14] | Daytime with haze, fair, light rain, and partially cloudy weather conditions far top | | University campus surveillance environment. Data gathered by mounting the camera on the rooftop of an 8-story building | 284 frames | Person Cars Poles | Raytheon 300D sensor core with 75 mm lens 30 Hz 320 x 240 |
| Four sets from KAIST Multispectral dataset [15] | Set 00 | Daytime Nearside | campus | 17,498 frames | Person Cars Poles Bicycles Bus | 20 Hz 640 x 480 |
| | Set 01 | Daytime Nearside | roadside | 8,035 frames | | |
| | Set 04 | Nighttime Nearside | roadside | 7,200 frames | | |
| | Set 05 | Nighttime nearside | downtown | 2,920 frames | | |

## A. Training and Learning Approach

In this work, we have included roadside objects for driver assistance comprising of seven different classes. It includes bicycles, bikes (motorbikes), buses, cars, dogs, pedestrians or people, and roadside poles as shown in Fig. 2.

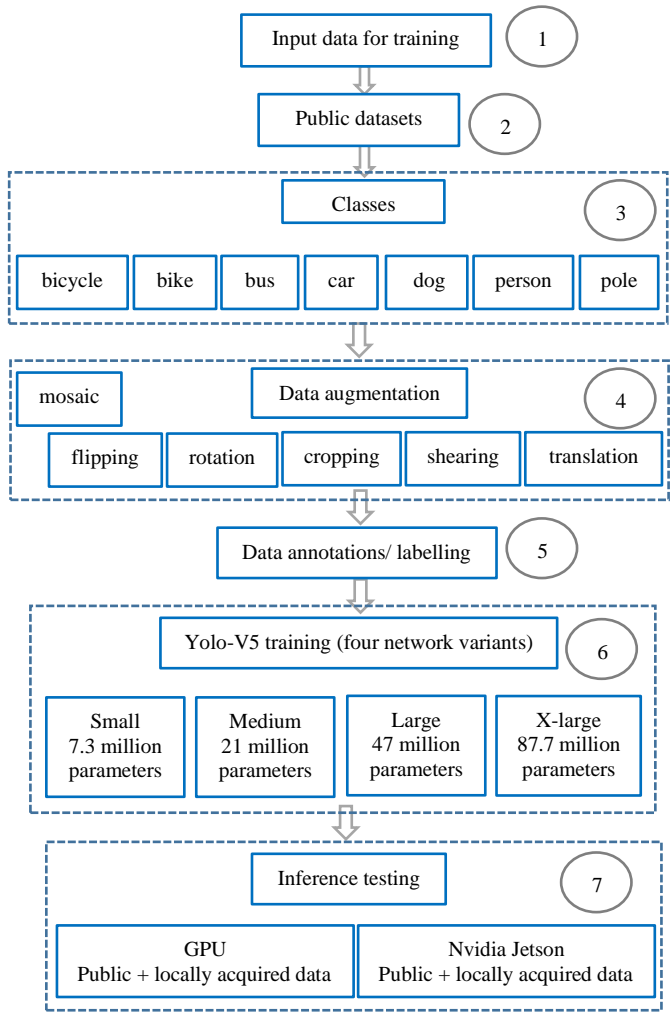

Fig. 2. Complete block diagram representation for object detection in thermal spectrum for driver assistance using end-to-end Yolo-V5 architecture.

The data samples from these classes are demonstrated in Fig. 3. These training data samples were gathered in different environmental conditions and were collected in the day as well as night-time.

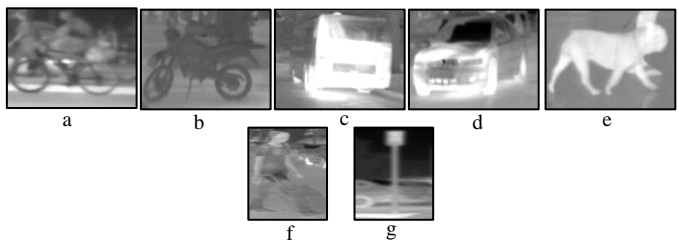

Fig. 3. Seven different data classes for training YOLO-V5 in thermal spectrum a) bicycle, b) bike, c) bus, d) car, e) dog, f) person, g) sign/ street pole.

The individual class-wise training data distribution is shown in Fig. 4. A total of 32,715 data samples has been used along with their respective class labels in the training process of the YOLO-V5 framework. Fig. 5 shows the data distribution of all the classes.

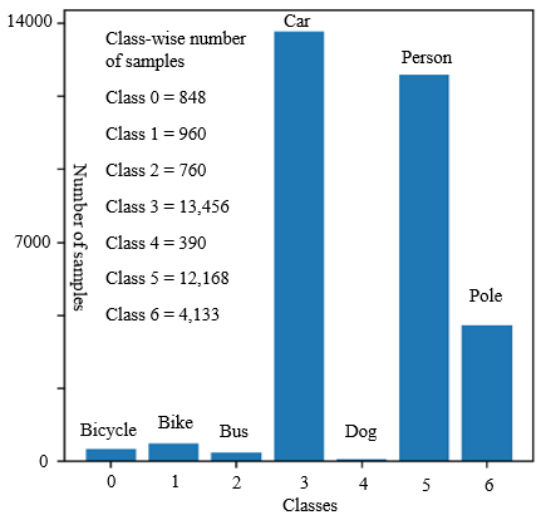

Fig. 4. Seven different data classes for training YOLO-V5 in thermal spectrum a) bicycle, b) bike, c) bus, d) car, e) dog, f) person, g) sign pole.

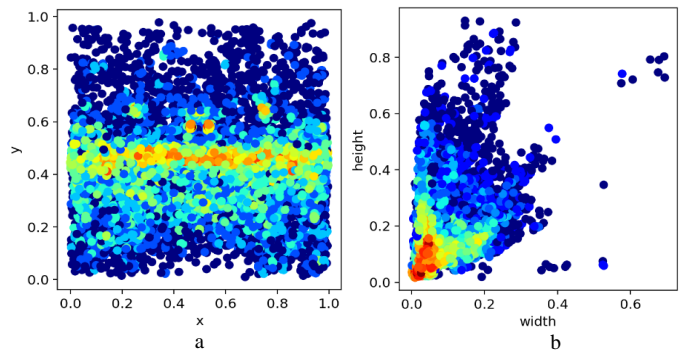

Fig. 5. Class-wise training data distribution a) un-distributed data samples, b) class-wise clustered training data samples.

In this work, we have focused on using Ultralytics [3, 4] resource for the training of YOLO-V5 architecture on our custom thermal dataset. During the training process, the configuration file is updated accordingly to our requirements specified in the head layer to adapt the number of classes (7 classes) on our dataset. To achieve precise training accuracy, we have trained all the network variants of YOLO-V5 architecture using both Stochastic Gradient Decent (SGD) [38] with momentum as well as Adaptive learning rate optimization (Adam) [39] optimizer. SGD is considered a state-of-the-art optimizer for training deep convolution neural networks. It works by performing parameter update for each training batch rather than updating the whole training batch at once. It performs faster on large training samples. Moreover, it is computationally less expensive and has the ability to converge much faster as compared to batch Gradient Decent (GD) optimizer [40]. Adaptive learning rate optimization (Adam) algorithm work by taking advantage of adaptive learning rates thus computing the individual learning rate for individual parameters. It has various key benefits in comparison to other state-of-art optimizers which include invariant to diagonal rescale of the gradients, it requires less amount of memory, it is computationally more resourceful and lastly, it is more appropriate for noisy gradients.

Lastly in this work rather than relying on one fixed learning rate we have used a one-cycle learning policy to find the optimal learning rate for our custom training set. The main reason for using this method is to achieve robust results during the training process of complex network variants of YOLO-V5 architecture. The algorithm works by following the Cyclical Learning Rate (CLR) to achieve faster training time with the regularization effect.

### B. Training Data Augmentation/ Transformation

Deep learning models are not considered ideal solutions with limited data options. To overcome this challenge, large data sets are required to perform optimal training of networks. Data augmentation is an effective way of producing a large number of new training samples with diversity using the existing datasets. The synthetically generated data samples can be used with the original data to build large training sets. In this work, we have used various data transformation methods as shown in block 4 of Fig. 2. The further details of each augmentation method are as follows.

- Flipping: This method is used to perform image flipping in different directions which include up, down, left, and right directions. It helps the model to be insensitive to subject orientation.
- Rotation: This method is used to add variability to rotations thus helping the model to be more resilient to the camera roll.
- Image cropping: This technique is used to augment changeability to positioning and size. It helps the model to be more resilient to subject translations and camera positions.
- Image shearing: It is used to add shifting to the image by providing desired vertical and horizontal angles.
- Translation: It is used to move the image along the horizontal and vertical axes. This method of transformation is very useful as objects can be located at almost anywhere in the image.
- Mosaic transformation: It is considered an advanced form of image augmentation operation. It works by combining different training samples in one image with varying ratios. We have employed this augmentation method during the training process. It helps the model to learn how to identify the objects at a smaller scale than normal. It also encourages the model to localize different types of images in different portions of the frame. Fig. 6 shows the mosaic transformation of four different training samples with different class labels.

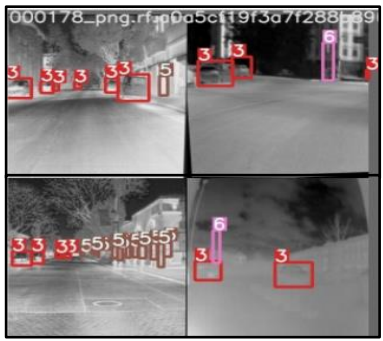

Fig. 6. Mosaic transformation formed by combining four different training samples with different class labels including cars labeled as 3, person labeled as 5 and poles labeled as 6.

### C. Data Annotations

In this study, we have performed manual bounding-based annotations for all the thermal classes. Tight bounding box-based annotations were performed on all the frames for the training of YOLO-V5 framework. All the network variants are trained to detect and classify bicycles, bikes, buses, cars, dogs, pedestrians, and roadside poles in stimulating environmental conditions which include sunny weather, cloudy weather, night-time with total darkness, daytime, and other challenging environmental conditions. Table II shows the respective distribution of all the annotated frames selected from four different public datasets in varying environmental conditions.

TABLE II
THERMAL DATA ANNOTATIONS

| Total frames selected from different public datasets | Total Annotated frames | Class wise annotations |
|---|---|---|
| 1. CVC-09: 1650 frames | 5100 frames with bounding box annotations | 1. Bicycles: 848 |
| 2. FLIR Adas: 1700 | | 2. Bikes: 960 |
| 3. KAIST Multispectral dataset: 1700 | | 3. Buses: 760 |
| | | 4. Cars: 13,456 |
| 4. OSU Thermal Dataset: 200 | | 5. Dogs: 390 |
| Total frames = 5,250 | | 6. Person: 12,168 |
| | | 7. Poles: 4,133 |
| | | Total annotations = 32,715 |

Fig. 7 shows some of annotated training data images in different environmental conditions and with different objects.

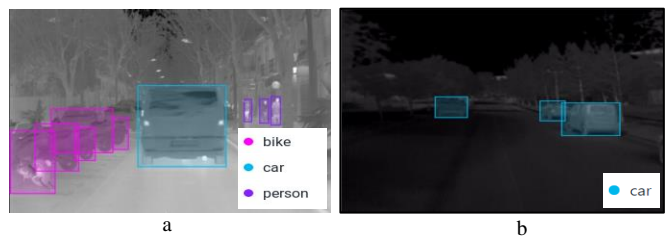

a                                          b

Fig. 7. Annotated sample images with tight bounding boxes from two different datasets a) CVC-09 dataset sample (road-side view), b) KAIST dataset sample (road-side parking).

### D. Locally Recorded Thermal Testing Data using LWIR Camera

The trained network variants are tested on both public as well as newly gathered test data to validate the efficacy of YOLO-V5 framework. The new test data is acquired using a camera based on a prototype uncooled micro-bolometer thermal camera array that embeds a Lynred [47] Long Wave Infrared (LWIR) sensor. Fig. 8 shows the images of the prototype thermal camera used in this research project [37]. Whereas Table III shows the technical specifications of the uncooled thermal camera. The data is collected in two different approaches. In, the first approach the data is gathered in a stationary manner by placing the camera at a fixed place. The camera is mounted on the tripod stand at a fixed height of nearly 30 inches such that the roadsides objects are covered in the video stream. The thermal video stream is recorded at 30 Frames Per Second (FPS). The data is recorded in different weather conditions.

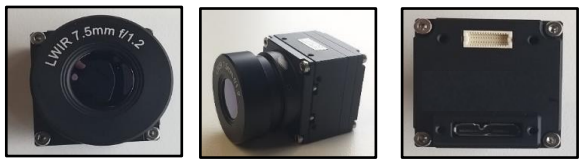

Fig. 8. Uncooled LWIR prototype thermal camera images from different angles developed under the Heliaus EU project [37].

TABLE III
TECHNICAL SPECIFICATIONS

| Prototype thermal camera specifications | |
|---|---|
| Quality and Type | VGA, Long Wave Infrared (LWIR) |
| Resolution | 640 x 480 pixels |
| Focal length (f) | 7.5 mm |
| F-Number | 1.2 |
| Pixel Pitch | 17 um |
| HFOV | 90-degree, 890 mm |

The data acquisition setup along with a complete road-side view is shown in Fig. 9.

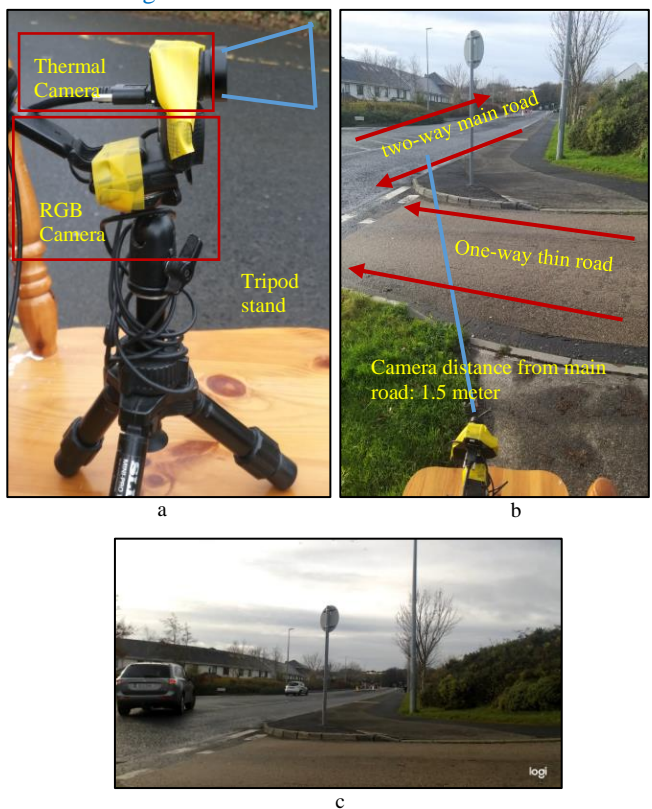

Fig. 9. Data acquisition setup by mounting the camera at a fixed place and height, a) thermal and RGB camera mounted at a height of 0.7 meters from the ground surface, b) roadside field of view, c) visual frame in the evening time with cloudy weather.

In the second approach rather than collecting data in a static approach the camera is mounted in the car and data is collected through the car. The main reason for collecting data in two different approaches is to check the effectiveness of trained network variants of YOLO-V5 framework on diversified and distinctive local data in different weather conditions. Fig. 10 shows the thermal camera along with visible camera setup on the car.

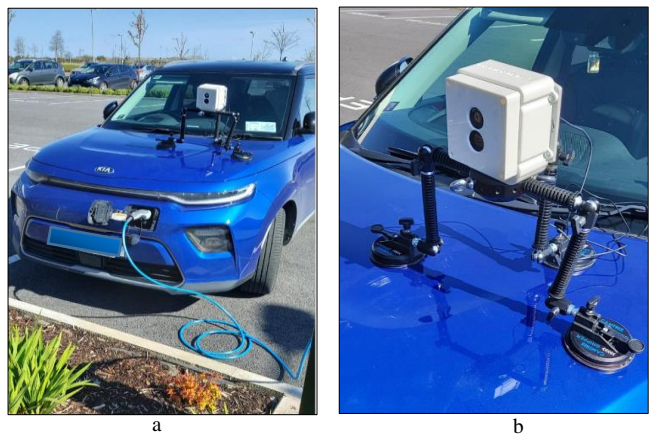

Fig. 10. Data acquisition setup by mounting the cameras on the car, a) thermal and RGB camera sealed in the white box and fixed on a suction tripod mount, b) closer view of the white box holding thermal and visible cameras.

Fig. 11 shows the recorded sample thermal frames in the day, evening, and night-time with different weather conditions using both static and by placing the camera on the car.

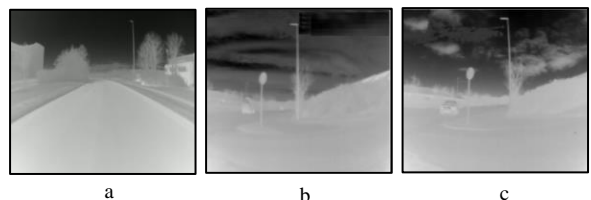

Fig. 11. Recorded thermal data samples in different weather conditions by placing the camera on the car and by placing the camera at a fixed place , a) day time with sunny weather, b) evening time with cloudy weather, c) night time with partially cloudy and windy weather.

IV. EXPERIMENTAL RESULTS ON GPU AND EDGE DEVICE

This section will exhibit the thermal object detection results along with the performance comparison of four different network variants of YOLO-V5 framework. In this study, different testing approaches have been employed thus making a fair numeric comparison between these approaches on the test dataset. These methods along with their experimental results are further discussed in subsections. Moreover, in this study, we have deployed the smallest model variant (in terms of having the least number of model parameters), and yet the fastest network variant (in terms of least inference time) of YOLO-V5 framework on Nvidia Jetson edge device [42]. This will eventually help us in the form of trained model portability for diversified ADAS applications.

In the initial phase of experimental results, we have used the pre-trained weights and tested them on thermal datasets without undergoing any training process. However, the results were not satisfactory as demonstrated in the ablation study (section VI) of this paper. In the next phase, we have trained all the networks of YOLO-V5 from scratch and used newly trained model weights for evaluation on thermal test datasets.

A. Training Configuration and Testing Approaches

This section will mainly focus on the training configuration using two different optimizers which include SGD and Adam employed in this study and different testing approaches. The complete training configuration is provided in Table. IV. The training process is performed on a server-grade machine equipped with XEON E5-1650 v4 3.60 GHz processor, 32 GB of ram, and GEFORCE RTX 2080 graphical processing unit. It comes with 12 GB of dedicated graphical memory, memory bandwidth of 616 GB/second, and 4352 cuda cores. During the training process, the training batch size is fixed to 4. The training process is performed on Pytorch deep learning framework [41]. It is important to mention that we have trained all the networks from scratch by un-freezing all the network layers and building new weights rather than transfer learning the networks to adapt the models for thermal data.

TABLE IV
YOLO-V5 TRAINING CONFIGURATION

| | Hyperparameter Selection | |
|---|---|---|
| 1 | Initial Learning Rate | 0.001 |
| 2 | Final learning Rate | 0.2 |
| 3 | Warmup initial bias learning rate | 0.1 |
| 4 | Learning policy | One cycle learning rate |
| 5 | Training and Testing Batch size | 8 & 32 |
| 6 | Weights | From scratch |
| 7 | Optimizer | SGD and Adam |
| 8 | Loss function | Binary Cross-Entropy (BCE) with Logits Loss |
| 9 | Epochs | 100 |
| 10 | Warmup epochs | 3 |
| 11 | Momentum | 0.9 |
| 12 | Warmup momentum | 0.8 |
| 13 | Weight decay | 0.0005 |
| 14 | IoU threshold | 0.5 |
| 15 | Anchor multiple threshold | 4.0 |

In the proposed study we have used three different test-time approaches which include test-time with no augmentation (TTNA), test-time augmentation (TTA), and test-time with model ensembling (TTME) methodology.

- TTNA: It is referred to as a conventional testing approach used for the unseen testing data provided to the trained object detection models. In this method, we don't perform any data augmentation/ transformation. Since no additional data augmentation operations are performed, the inference time depends on how dense the trained model is.

- TTA: Test-time augmentation is often helpful to achieve more robust results in the form of high inference accuracy from trained networks. Test-time augmentation is an extensive application of data augmentation applied to the test dataset. Specifically, it works by creating multiple augmented copies of each image in the test set, having the model make a prediction for each, then returning an ensemble of those predictions. However, since the test dataset is enlarged with artificially augmented images the inference time also increases as compared to NA which is one of the drawbacks of this approach. In this study, test-time augmentation method is performed on the test dataset by incorporating three different augmentation methods which include image shifting, cropping, and flipping.

- TTME: This is a technique for establishing the performance of multi-modal trained network variants on the test datasets. In machine learning model ensembling or ensemble learning refers to as using multiple trained networks at the same time in a parallel manner to produce one optimal predictive inference model. In this study, we have tested the performance of individually trained variants of the Yolo-V5 framework and selected the best combination of models which in turn helps in achieving better mean-average precision (mAP) scores on the validation set. However, as the trade-off, the individual inference time on each test frame/ image increases relatively as compared to NA methods.

*B. Training Results*

This section will summarize the training results of all the network variants of YOLO-V5 framework. The training accuracy and loss results are analyzed using different quantitative metrics to fully evaluate the effectiveness of all the trained models. The overall loss in the YOLO-V5 framework is calculated as compound lost based on three different scores which include objectness score, class probability score, and bounding box regression score. In this study, we have used Binary Cross-Entropy (BCE) with Logits Loss function in pytorch for loss calculation of class probability and object score. Whereas, the model accuracy is computed in terms of recall rate, model precision, and mean average precision (mAP). These accuracy metrics are explained below respectively.

*1) Recall and Precision*

In machine learning recall or sensitivity is counted as a critical statistical tool which is also referred to as true positive rate. It is defined as the ratio of true positive and the total amount of ground truth positives. The precision of any class is defined as the ratio of true positive (TP) and the sum of predicted positives. It is also referred to as positive predicted values. Equation (1) shows the formula of recall and precision metrics.

$$Recall = \frac{tp}{tp+fn} \text{ X } 100 \quad Precision = \frac{tp}{tp+fp} \text{ X } 100 \quad (1)$$

Where tp is the true positives, fn is defined as false negatives and fp is the false positives.

*2) Mean Average Precision (mAP):*

The mean average precision (mAP) is a standard metric used to measure the performance of deep learning models trained for applications such as information retrieval and object detection tasks. It is defined as the area under the Precision-Recall curve. The mAP for the object detection model is the average of the AP computed for all the classes. Equation (2) shows the formula for calculating the AP.

$$AP = \sum_{Recall_i} Precision\ (Recall_i) = 1 \quad (2)$$

As mentioned earlier in (Section III-A), we have used both SGD and ADAM optimizers during the training process and selected the trained models with the best performance for validation on public as well as locally gathered test data. Table. V shows the area under the Precision-Recall curve for all the classes of four different network variants trained from scratch using both SGD and ADAM optimizer. Also, during the training process, we have made a comparative analysis of total graphical memory usage and total training time required for all the models. For a better understanding of the performance comparison of all the models, Table. VI shows the numerical results of all the accuracy metrics, loss metrics, graphical memory usage, and overall training time required using both SGD and ADAM optimizer.

TABLE V
YOLO-V5 TRAINING RESULTS

| Model | Optimizer: SGD | Optimizer: Adam |
|---|---|---|
| | Precision-Recall Curve of all four network variants of YOLO-V5 | |
| Small | mAP = 90.43% | mAP = 89.80% |
| Medium | mAP = 91.26% | mAP = 91.05% |
| Large | mAP = 90.53% | mAP = 89.57% |
| X-large | mAP = 91.31% | mAP = 90.30% |

It can be observed and summarized from Table. V and Table. VI that models trained using SGD optimizer have performed significantly better as compared to models trained using ADAM optimizer in the terms of better mAP value, better precision scores, lower GPU usage, lower training time, and last but not least lower losses. Also, by analyzing the individual performance of all the models, the X-large model has achieved the best mAP score of 91.31% with the lowest losses as compared to all other models. Whereas the medium network variant has scored the second-best mAP score of 91.26% along with the highest recall rate of 94.51% among all the models. In terms of the highest precision scores, the large and x-large model has outperformed all other models thus achieving the best precision score of 75%. Fig. 12 shows the accuracy and loss graphs of the X-large model as it has achieved exceptional performance in terms of the highest mAP scores and lower loss values when analyzing the performance of other network variants of YOLO-V5 framework. However, as the trade-off, this model requires the highest training time and greater GPU usage which makes it computationally more expensive. The overall model is consisting of 407 layers, 88.47 million gradients, and the final trained model weight size is 173 MB.

TABLE VI
YOLO-V5 TRAINING PERFORMANCE COMPARISON

| Optimizer | GPU usage/ epoch (GB) | Training time required (Hours) | P % | R % | Losses | | |
|---|---|---|---|---|---|---|---|
| | | | | | Bounding box loss | Objectness loss | Classification loss |
| Model: Small | | | | | | | |
| SGD | 2.53 | 0.8 | 69.63 | 93.83 | 0.026 | 0.025 | 0.00079 |
| ADAM | 2.58 | 0.708 | 61.92 | 94 | 0.028 | 0.027 | 0.00095 |
| Medium | | | | | | | |
| SGD | 3.11 | 1.2 | 70.84 | 94.51 | 0.023 | 0.022 | 0.00070 |
| ADAM | 3.34 | 1.24 | 67.68 | 94.23 | 0.027 | 0.025 | 0.00115 |
| Large | | | | | | | |
| SGD | 6.65 | 1.8 | 75 | 93 | 0.021 | 0.020 | 0.00058 |
| ADAM | 6.68 | 1.83 | 68.92 | 92.51 | 0.025 | 0.025 | 0.00112 |
| X-large | | | | | | | |
| SGD | 9.74 | 3.2 | 75 | 92.89 | 0.020 | 0.019 | 0.00062 |
| ADAM | 9.82 | 3.25 | 69 | 93.76 | 0.024 | 0.023 | 0.00117 |

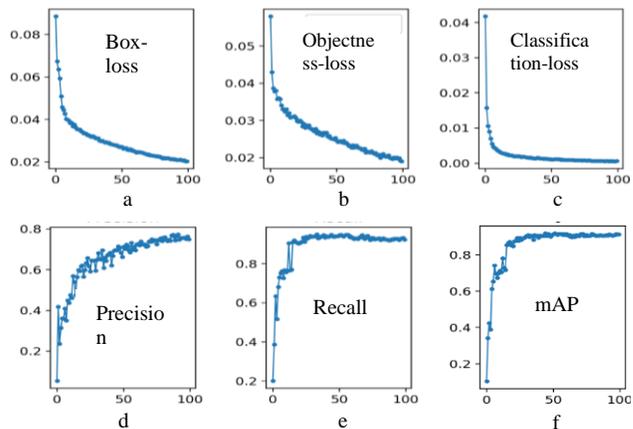

Fig. 12. Training and Loss graphs of X-large model using SGD optimizer a) bounding-box loss, b) Objectness loss, c) Classification loss, d) Model Precision, e) Model recall curve, and f) Mean average precision (mAP).

*C. Validation Results on Public and Locally Gathered Test-Data*

In the first phase, the performance of YOLO-V5 trained networks is evaluated on the unseen data gathered from publicly available datasets. As discussed earlier in (Section IV-A) three different test approaches are used which include TTNA, TTA, and TTME to validate the efficacy of four different networks of YOLO-V5 architecture. The training results signify that the SGD optimizer has performed better than the ADAM optimizer, however, during the testing phase, we have included the results extracted from trained networks using ADAM optimizer thus making fair testing evaluation among all the set of trained models. The test data comprises different weather conditions, different environments/ places, and varying distances of the objects from the camera. Table. VII shows the total number of frames used as the test data from four different publicly available datasets along with their respective attributes.

TABLE VII
TEST DATA COLLECTED FROM PUBLIC DATASETS

| Public Datasets | Weather Conditions | Environment | Frames selected | Total frames |
|---|---|---|---|---|
| 1 CVC-09 | Daytime and nighttime | Roadside | 1000 + 1000 | 2436 |
| 2 FLIR-ADAS | Daytime and nighttime with Sun and cloudy weather condition | Roadside | 250 | |
| 3 KAIST Multispectral | Nighttime | Downtown | 136 | |
| 4 OSU Thermal Dataset | Daytime with cloudy weather conditions | University campus environment | 50 | |

In the first segment, the inference test is run on test data using the TTNA approach. Whereas, in the second part we have run the inference results using the TTA approach. During the complete testing phase, the confidence threshold is set to 0.5. Fig. 13 shows the sample results on sixteen different frames selected arbitrarily from the test-set using the TTNA approach and models trained using the SGD optimizer. These frames consist of either single or multiple objects in the thermal spectrum from seven different classes as shown in Fig. 3. The test results are sub-divided into four parts extracted from four network variants of the YOLO-V5 framework.

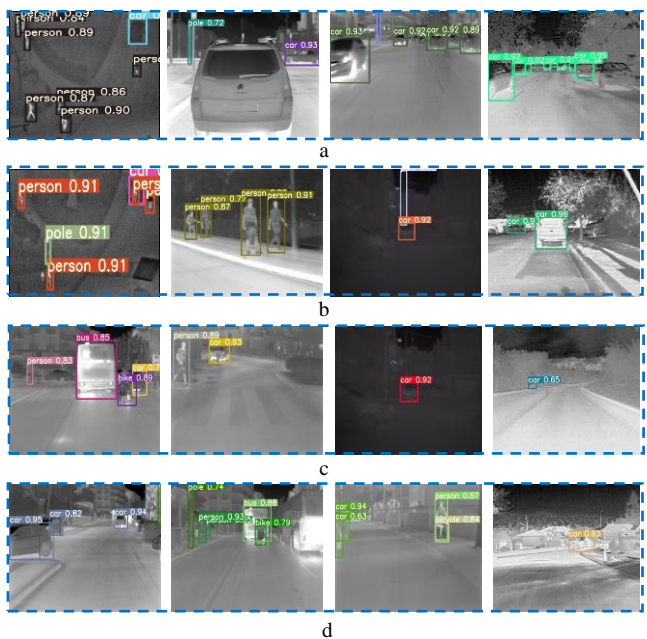

Fig. 13. Object detection inference results with class confidence scores on sixteen different frames from four different public datasets using test-time without augmentation approach (TTNA) a) results extracted using small network variant, b) results extracted using medium network variant, c) results extracted using large network variant and d) results extracted using x-large network variant.

It can be observed from Fig. 13 that inference results on test data using different network variants trained on thermal data have improved significantly as compared to results shown in abalation study and Fig. 23 which were generated using pre-trained weights. However, by closely analyzing the results still, we are unable to detect and classify some of the objects in thermal frames with challenges like scale and view-point variations, occlusions, and overlapping classes. For instance, in Fig. 13a frame 2 we are unable to detect the car close to camera mounted on another car for recording the data. Similarly, in Fig.13b frame 1 rather than detecting two cars, the medium model can detect only one car.

To overcome these issues and further improve the test accuracy, the inference test is run using the test-time augmentation (TTA) approach. The average inference time per frame using the TTNA method varies depending on the size of the model. The average inference time using the small model is 11 milliseconds whereas the average inference time using the x-large model is 21 milliseconds. Fig. 14 depicts the inference results on eight different samples from the test-data using TTA approach and four different network variants of the YOLO-V5 framework using SGD optimizer. It can be observed that results are improved marginally as compared to the TTNA approach and significantly as compared to using originally pre-trained weight shown in Fig. 23. This technique helps in detecting and classifying the object in the thermal spectrum more robustly with data complications such as occlusion, overlapping classes, scale variation, and varying environmental conditions. However, as the trade-off the average inference time per frame by employing TTA method increases as compared to TTNA method since additional augmentation operations are performed during the testing phase.

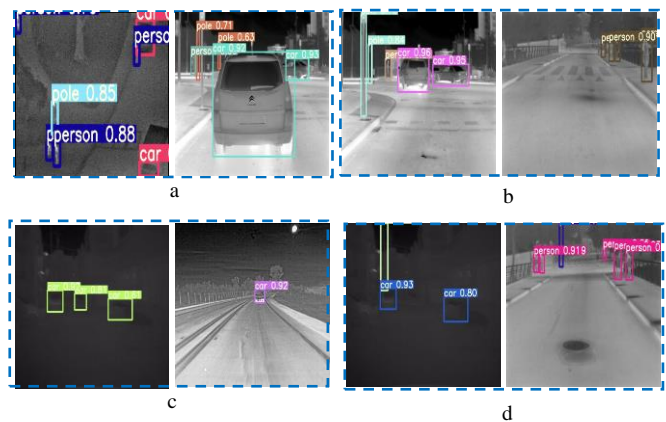

Fig. 14. Object detection inference results with class confidence scores on eight different frames from four different public datasets using test-time with augmentation approach (TTA) a) results extracted using small network variant, b) results extracted using medium network variant, c) results extracted using large network variant and d) results extracted using x-large network variant.

Table. VIII shows the numerical performance comparison between TTNA and TTA for all the network variants by computing the mAP, precision, recall rate, and average inference time required per image. For this purpose, we have short-listed a test-set of 250 frames with image complexities like occlusions, scale and viewpoint variations, and especially multiple objects with closely overlapping classes from the overall test-data as shown. By analyzing the results from Table. VIII we can summarize that the large model using TTA method and SGD optimizer has achieved the best mean average precision, recall rate, and precision score of 86.6%, 90.2%, and 86.5% respectively as compared to other network variants of YOLO-V5 framework.

TABLE VIII
YOLO PERFORMANCE EVALUATION ON TEST IMAGES FROM PUBLIC DATASETS (the best value per metric is highlighted in green and emboldened)

| Optimizer | Method Test time with no Augmentation (TTNA) | | | | Method Test time Augmentation (TTA) | | | |
|---|---|---|---|---|---|---|---|---|
| | Precision | Recall % | mAP % | Average inference time/ image (millisecond) | Precision | Recall % | mAP % | Average inference time/ image (millisecond) |
| Model: Small | | | | | | | | |
| SGD | 88.4 | 82.5 | 79.5 | 11 | 83.5 | 89.8 | 85.4 | 23 |
| ADAM | 85.9 | 81.5 | 77.8 | 11 | 72.9 | 85.8 | 81.8 | 21 |
| Model: Medium | | | | | | | | |
| SGD | **90.5** | 86.5 | 83.2 | 13 | 86.2 | **90.1** | 85.8 | 28 |
| ADAM | 86.5 | 81.4 | 78.3 | 13 | 82 | 87.5 | 83.7 | 28 |
| Model: Large | | | | | | | | |
| SGD | 89.3 | **87.2** | **84.1** | 17 | **86.5** | 90.2 | **86.6** | 35 |
| ADAM | 84.7 | 81,6 | 78.1 | 18 | 80.5 | 87.8 | 83.3 | 37 |
| Model: X-Large | | | | | | | | |
| SGD | 88.8 | 85.6 | 82 | 21 | 85.7 | 89.3 | 85.2 | 53 |
| ADAM | 86.2 | 76 | 72 | 23 | 80.2 | 86.7 | 81.5 | 54 |

To further enhance the testing accuracy and explicitly reduce the inference time as compared to the TTA method a third testing approach i.e. model ensembling is used in this study. In this approach, we have tried various combinations of models by running them in ensembling style and selected a set of two best models by evaluating their performance on test data as shown in Fig. 15.

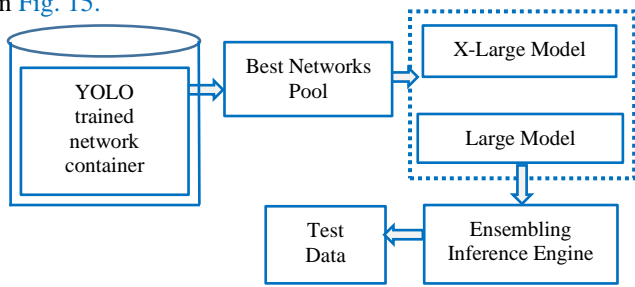

Fig. 15. Model ensembling inference engine architecture.

As demonstrated in Fig. 15 the ensembled inference engine is consists of large and x-large models to produce one optimal predictive model. It is then evaluated on the test data as shown in Table. VII. Fig. 16 shows the individual inference results on four selected frames with complex scenarios like multiple objects with overlapping classes, object scale and viewpoint variations, and different weather conditions. Whereas Table. IX shows the numerical performance in terms of mAP, precision, recall rate, and average inference time required per image using the model ensembling method.

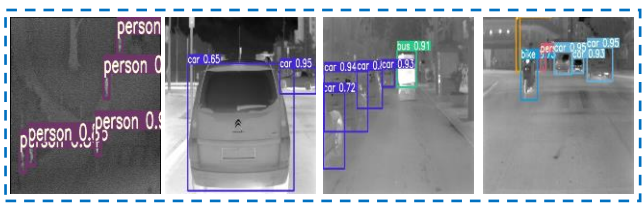

Fig. 16. Model ensembling inference results on four different frames with multiple objects, overlapping classes, varying distance of the object from the camera, and different environmental and weather conditions.

TABLE IX
MODEL ENSEMBLING PERFORMANCE EVALUATION ON TEST IMAGES FROM PUBLIC DATASETS

| Optimizer | Method Test time with Model Ensembling (TTME) | | | |
|---|---|---|---|---|
| | Precision % | Recall % | mAP % | Average inference time required/ image (millisecond) |
| SGD | 87.6 | 88.8 | 85.5 | 33 |

In the second phase, we have used the same model ensembling approach to validate its robustness on locally gathered test data. As mentioned earlier in (Section III-D), test data is collected in two different methods which include mounting the camera at a fixed place, and in the second method, data is gathered through the car. The locally gathered test data comprises different weather conditions, different environments/ places, multiple objects with class and scale variations. Table. X shows the total number of short-listed frames used as the test data from our locally generated dataset along with their respective attributes.

TABLE X
LOCALLY GATHERED TEST DATA

| Method | Weather Conditions | Environment | Frames selected | Total frames |
|---|---|---|---|---|
| By mounting the camera at a fixed place | Daytime, evening time, and nighttime with cloudy and windy weather | Roadside | 5000 | 20,000 |
| By mounting the camera on the car | Daytime and evening time with sunny and cloudy weather | Cityside and University Campus | 15000 | |

Fig. 17 displays the inference results on eight different frames gathered from the uncooled prototype LWIR thermal camera used in this project. It can be observed that an ensemble inference engine comprising of x-large and large network variant has achieved precise results as it can detect and classify multiple objects of different classes in newly gathered local test data.

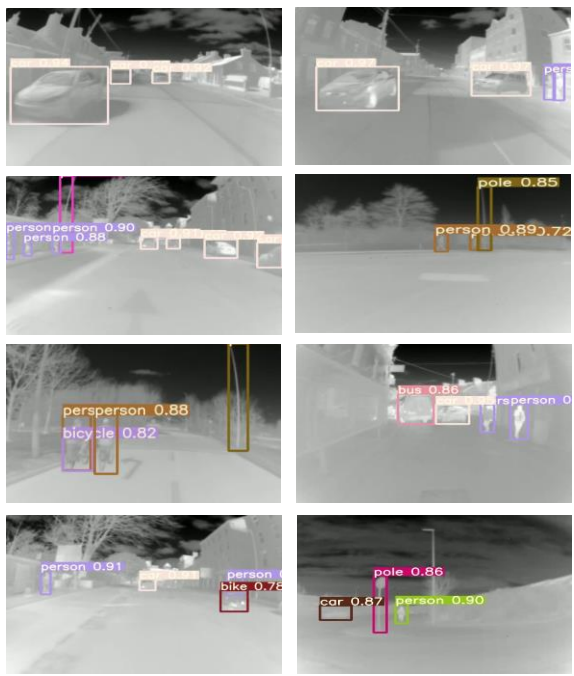

Fig. 17. Model ensembling inference results on eight different frames acquired from prototype thermal camera with multiple objects, overlapping classes, the varying distance of the object from the camera, and different environmental and weather conditions.

However, to further investigate its effectiveness, we have computed various accuracy metrics which include precision, recall, mAP, and mean inference time required per image as it was computed in the case of public datasets on a set of 250 thermal frames. These results are shown in Table. XI.

TABLE XI
MODEL ENSEMBLING PERFORMANCE EVALUATION ON TEST IMAGES FROM THE LOCALLY GATHERED DATASET

| Optimizer | Method Test time with Model Ensembling (TTME) | | | |
| --- | --- | --- | --- | --- |
| | Precision % | Recall % | mAP % | Average inference time required/ frame (millisecond) |
| SGD | 83.5 | 78.1 | 70 | 29 |

### D. Deployment and Validation Results on Edge Computing

After successful convergence and testing of YOLO-V5 networks on GPU architecture, in the next step, we drive towards the deployment of the trained network on edge-inference architecture. The primary goal is to create a flexible, scalable, secure, and more automated hardware system thus allowing us to easily export the trained network weights for easy model portability. The core benefits of deploying the trained machine learning (ML) model on edge devices include.

1. The edge hardware is assumed to be more energy-efficient since it requires less amount of power resources as compared to single or clustered-based CPU and GPU server machines.
2. Locating and processing inference at the edge architecture lies in saving communication power.
3. Last but not least the cost of edge-based inference hardware is considerably less as compared to other computational hardware such as field-programmable gate arrays (FPGA) and GPUs.

For this study, Nvidia Jetson nano [42] developer kit is selected to evaluate the performance of YOLO-V5 trained model on thermal test data. It is a small yet powerful edge computer that allows us to run multip le neural networks in parallel for several computer vision applications such as image classification, object detection, segmentation, and speech processing. It is considered as all in an easy-to-use platform that runs in as little as five watts of power. Fig. 18 shows the Nvidia Jetson nano developer kit equipped with CPU QUAD-core ARM A57 at 1.43 GHz and GPU 128-core Maxwell. It comes with the memory of 4 GB, 64-bit, LPDDR4 25.6 GB/s. Jetson nano has a total of four USB 3.0 ports, HDMI port, an ethernet port, and barrel connector to power it via five volts and 4 ampere supply.

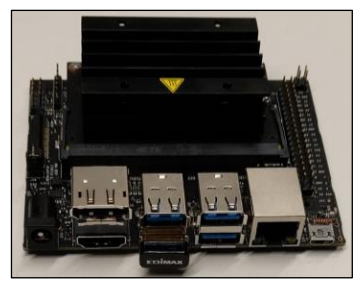

Fig. 18. Nvidia Jetson nano developer kit for deploying the YOLO-V5 trained networks.

The small network variant of the YOLO-V5 framework is selected as it has the minimum number of model parameters and requires the least inference time as compared to other models of YOLO-V5 which makes it a computationally less expensive and cost-effective model. For better optimization, and further reduce the inference time we have used TensorRT inference optimizer [43]. It is a type of deep learning inference optimizer and runtime engine that delivers low latency and high throughput for deep learning inference applications. TensorRT-based inference engines can perform up to 40X faster than CPU-only platforms. The optimized inference models can be easily deployed to hyper-scale data centers, embedded and edge devices, and automotive product platforms. Fig. 19 shows the structural architecture design for converting the YOLO-V5 small deep learning model trained on thermal data to an optimized inference engine. As shown in Fig. 19 the process overflow for converting the trained network variant of the YOLO-V5 framework to TensorRT based optimized inference engine is consists of six main steps. In the first step, it maximizes throughput by quantizing models to 8-bit integer data type while preserving the accuracy. In the second step, it improves the use of GPU memory and bandwidth by fusing nodes in a kernel. In the next step, it performs Kernal auto-tuning. In the fourth step, it minimizes memory footprints and re-uses memory for tensors efficiently. In the last steps, it processes multiple input streams in parallel and finally optimizes neural networks periodically with dynamically generated kernels [43]. Once the model is optimized successfully, it is serialized and deserialized to run the inference test. Fig. 20 shows the inference results on ten different thermal frames from the public as well as locally gathered test data. The generated results are in the form of bounding boxes with the respective class number.

Fig. 21 shows the test data results on 400 images in the form precision-recall curve for all the classes from both local and public test data on Jetson nano.

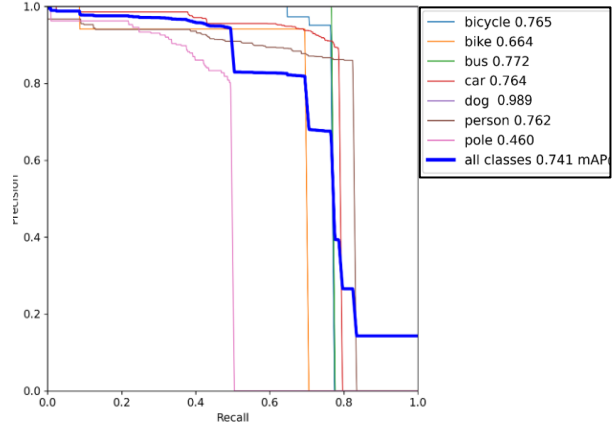

Fig. 21. Test data results for all the classes on Jetson Nano using small network variant with an overall mAP of 74.1%.

Table. XII shows the comparative analysis of the inference time per frame and FPS rate between the typically trained model tested on GPU, optimized/ accelerated version of the model tested on GPU, and accelerated version of the model tested on Nvidia Jetson nano.

TABLE XII
INFERENCE TIME AND FPS COMPARISON

| Model: YOLO-V5 Small Variant | | |
|---|---|---|
| Average inference time (milliseconds) /image and frames per second (FPS) comparison chart on GPU and Jetson Nano | | |
| Non-optimized version on GPU | Optimized version on GPU | Optimized version on Nvidia Jetson Nano |
| 11 ms | 5 ms | 320 ms |
| ≈ 90 FPS | ≈ 170 FPS | ≈ 3 FPS |

It can be observed from the Table XII that inference time has reduced to nearly 45% on GPU by using the optimized version of YOLO-V5 model through TensorRT which will eventually benefit us when running the inference test with a large number of test frames and subsequently running the inferences on higher frames per second (FPS) videos. Whereas on Nvidia Jetson it requires nearly 320 millisecond average inference time per frame and FPS rate of 3 with image resolution of 640x480 pixels.

V. DISCUSSIONS/ ANALYSIS

This section will mainly emphasize on individual training and testing performance comparison of all the model variants of YOLO-V5 framework.

- The small variant requires the lowest inference time among all other models with the mAP score of 85.4% using TTA approach which is nearly equal to the mAP score of the model ensembling method that is 85.5% during the testing phase.
- The medium model tends to achieve the best precision score of 90.5% using TTNA method and the best recall score of 90.1% using the TTA method among all other models during the testing phase. However, this model was unable to achieve robust mAP scores during both the training and testing phases.
- The large variant proves to the best network by achieving the highest mAP scores using both TTNA and TTA methods of 84.1% and 86.6% respectively during the testing phase as compared to other trained networks. However, it requires a longer inference time specifically when using the models with TTA method which is nearly 35 millisecond per frame.
- The X-large model turns out to be the best-trained model thus scoring the highest mAP score of 91.31% and lowest losses using the SGD optimizer but during the testing phase, the model was unable to achieve exceptional accuracy on the validation/ test set along with the highest inference time using both TTNA and TTA methods. However, this model comes up with the best possible match with large network variant in the model ensembling method.
- By examining the overall performance of all the models, we can conclude that test accuracy using the TTA approach is significantly better as compared to TTNA. However, as a trade-off, the TTA method requires huge inference time which makes it a computationally more expensive method.
- Fig. 22 shows the maximum and minimum inference time comparison chart of all the trained models tested on GPU using three different testing approaches. It can be observed that small network variant requires the least inference time using both TTA and TTNA methods thus making this model computably the least expensive and ideal network for real-time deployments especially on edge devices with comparatively less computational power. Also, the minimum and maximum time required by TTME approach is smaller as compared to TTA method which makes it more time-efficient networks.

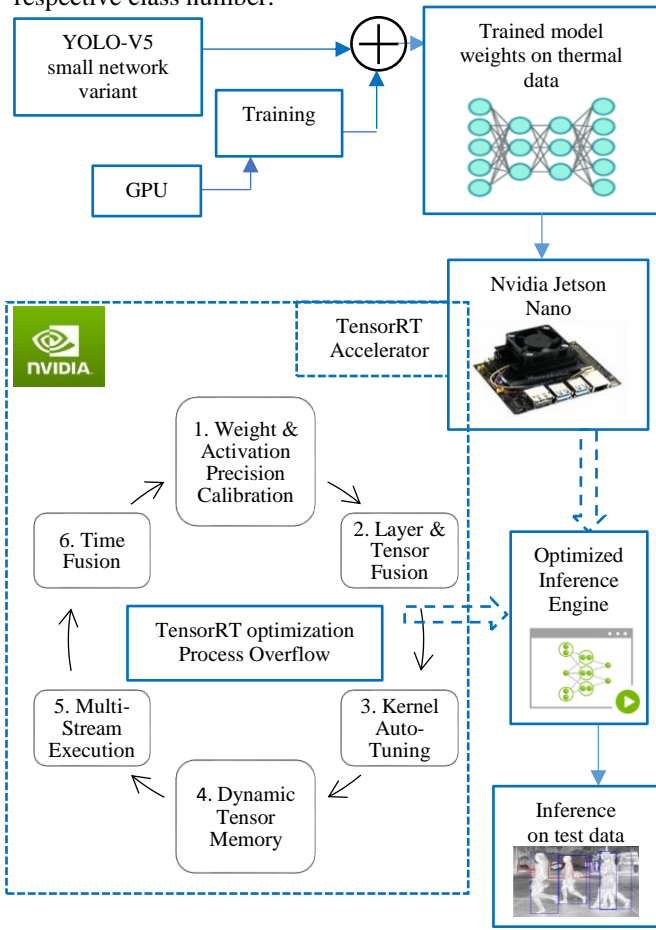

Fig. 19. Structural architecture for converting YOLO-V5 trained network to TensorRT optimized inference engine.

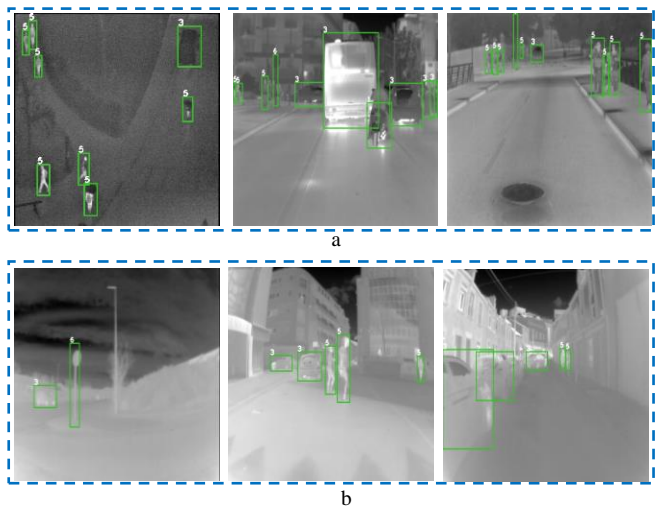

Fig. 20. Inference results on Nvidia Jetson nano using TensorRT optimization engine by showing respective class numbers, a) inference results on public datasets, b) inference results on the locally gathered test dataset.

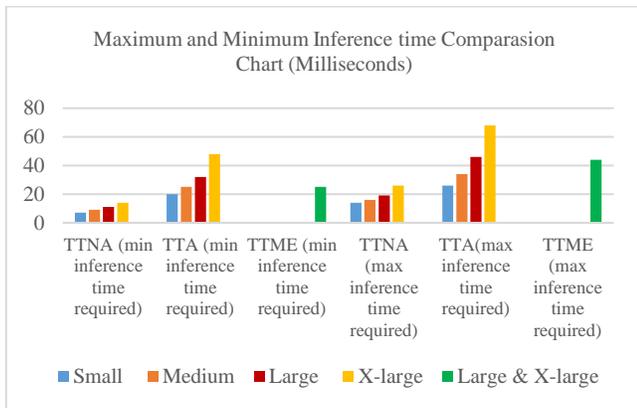

Fig. 22. Inference time comparison chart of the trained models on GPU using test-time with no augmentation, test-time with augmentation, and test-time with model ensembling methods.

Lastly, the newly trained model weights performed much better as compared to pre-trained weights however still in some most complex frames, models seem to provide inadequate results which are discussed and shown in some of the cases as follows.

Case 1

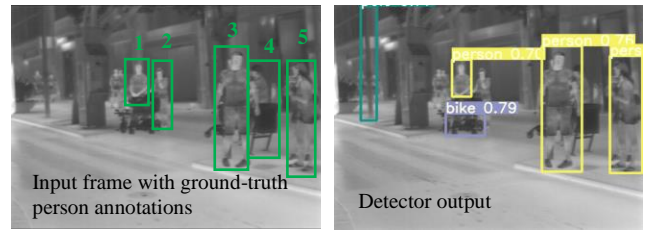

Case 2

These results are obtained using the small network variant using TTNA method. It can be observed from the left side input frame that five people can be seen from a human perspective marked with manually annotated green boxes for better understanding. The right-side frame shows the detector output. The network was only able to detect three people since the second person was putting a hand in front of her face whereas the fourth person view was a side pose with occluded vision thus detector was unable to detect the second and fourth person. Moreover, we can see a second person holding a baby walker however the detector miss-classified it as a bike.

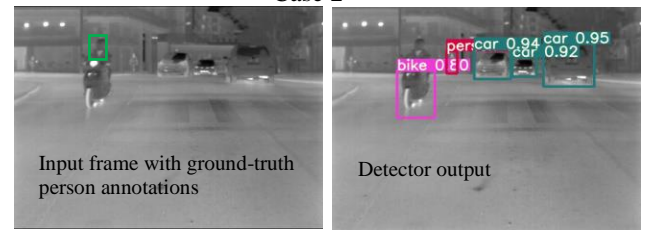

This result is obtained using the large variant using TTA method. There is a total of five objects in this frame (three cars and two people). The detector was able to detect four objects with good confidence scores however, it was failed in detecting one other person as demonstrated in the left side frame. This is because the person is riding the bike with occluded vision (wearing the helmet) which makes his personal features nearly blurred thus the model fails to detect and classify it.

## VI. ABALATION STUDY

This section shows an ablation study by analyzing the results using the pre-trained weights of all the network variants and testing them on thermal datasets without undergoing any training process. The results on the OSU-Thermal public thermal dataset [14] are demonstrated in Fig. 23 using different pre-trained network variants of the YOLO-V5 framework. The output results were not satisfactory as we were unable to detect most of the objects along with zero and wrong predictions in the thermal frames. A similar type of results was observed in the case of all the public and locally acquired test sets.

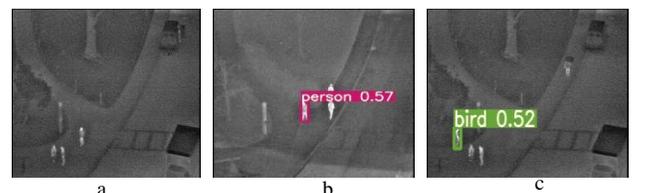

Fig. 23. Unsatisfactory thermal object detection results on three different frames from OST-Thermal dataset using initial pre-trained weights, a) no object detected using all the network variants, b) 1 person detected out of 3 persons and 2 vehicles using large network, c) person misclassified as bird using medium network whereas other objects are not detected.

## VII. CONCLUSION/ FUTURE WORK

In this work, we have proposed an AI-based thermal imaging object detection pipeline for ADAS application. We have employed four different network variants of the YOLO-V5 framework and trained them using four public datasets using SGD as well as ADAM optimizer. The X-Large model turns out to be the best-trained model thus achieving the highest mean average precision. The performance estimation of trained network variants is validated using both public as well as locally gathered new test data in different weather and environmental conditions. The Large network variant comprising 47.4 million parameters has achieved the best mAP score of 84.1% using TTNA and 86.6% using the TTA method. To further reduce the inference time as compared to the TTA method without compromising the accuracy we have used the TTME method. X-large and large model proves to be the best network coupler thus producing the results as one optimal inference engine. The model ensembling-based inference engine achieves the overall mAP of 85.5% and 70% on public and locally gathered test data respectively. Secondly, we have used TensorRT optimizer to further reduce the model inference time that can eventually help in real-time deployments, especially on edge devices. The optimized version of the trained model is tested on both GPU as well as Jetson nano. However, at the current stage, the inference results on edge device even with deploying the smallest variant of YOLO-V5 is not too much appealing for real-time ADAS implementations. It is evidenced by the fact that the ratio of FPS rate between Jetson nano and GPU is 3:170. Also, the average inference time per frame on Jetson nano is nearly 98% more as compared to inference time on GPU.

As the possible future directions, these systems can be deployed on more powerful edge devices with a higher flop rate and less operating power for optimal performance, especially in real-time environments. Moreover, we intend to include more thermal classes thus making the overall system more mature and robust. In addition to this, we can further integrate the current object detection system with object tracking thus to estimate the position of an object, as well as incorporate position predicted by dynamics. This will eventually help us in counting the number of vehicles, pedestrians, etc, along with their approximate estimated distance. One such example is demonstrated in Fig. 24 where we have integrated deep association metrics [44] tracking with YOLO-V5 for person tracking using our locally gathered test data.

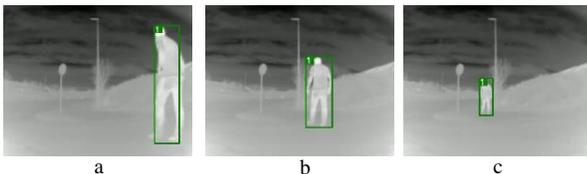

Fig. 24. Inference results along with person tracking by assigning id 1 a) frame 1, b) frame 15, c) frame 23.

## REFERENCES

[1] Kala, Rahul. "Advance Driver Assistance Systems," in *On-road intelligent vehicles: Motion planning for intelligent transportation systems*. Butterworth-Heinemann, 2016, pp. 59–80.

[2] Munir, Farzeen, et al. "Thermal Object Detection using Domain Adaptation through Style Consistency." *arXiv preprint arXiv:2006.00821* 2020.

[3] YOLO-V5 GitHub repository, [Online] Available: https://github.com/ultralytics/yolov5, (Last accessed on 26th February 2021).

[4] YOLO-V5 Website resource, [Online] Available: https://zenodo.org/badge/latestdoi/264818686 , (Last accessed on 26th February 2021).

[5] J. Redmon, S. Divvala, R. Girshick, and A. Farhadi, "You only look once: Unified, real-time object detection," in *Proceedings of the IEEE Computer Society Conference on Computer Vision and Pattern Recognition*, 2016, vol. 2016-December, DOI: 10.1109/CVPR.2016.91.

[6] Redmon, Joseph. "Darknet: Open source neural networks in c.", 2013.

[7] W. Liu *et al.*, "SSD: Single shot multibox detector," in *Lecture Notes in Computer Science (including subseries Lecture Notes in Artificial Intelligence and Lecture Notes in Bioinformatics)*, 2016, vol. 9905 LNCS, doi: 10.1007/978-3-319-46448-0_2.

[8] Girshick, Ross, Jeff Donahue, Trevor Darrell, and Jitendra Malik. "Rich feature hierarchies for accurate object detection and semantic segmentation." in *Proceedings of the IEEE conference on computer vision and pattern recognition*, 2014, pp. 580-587.

[9] Girshick, Ross. "Fast r-cnn." in *Proceedings of the IEEE international conference on computer vision*, 2015, pp. 1440-1448.

[10] K. He, G. Gkioxari, P. Dollár, and R. Girshick, "Mask R-CNN," *IEEE Trans. Pattern Anal. Mach. Intell.*, vol. 42, no. 2, 2020, DOI: 10.1109/TPAMI.2018.2844175.

[11] T. Y. Lin *et al.*, "Microsoft COCO: Common objects in context," in *Lecture Notes in Computer Science (including subseries Lecture Notes in Artificial Intelligence and Lecture Notes in Bioinformatics)*, 2014, vol. 8693 LNCS, no. PART 5, DOI: 10.1007/978-3-319-10602-1_48.

[12] Deng, Jia, Wei Dong, Richard Socher, Li-Jia Li, Kai Li, and Li Fei-Fei. "Imagenet: A large-scale hierarchical image database." in *2009 IEEE conference on computer vision and pattern recognition*, Ieee, 2009, pp. 248-255.

[13] M. Everingham, L. Van Gool, C. K. I. Williams, J. Winn, and A. Zisserman, "The pascal visual object classes (VOC) challenge," *Int. J. Comput. Vis.*, vol. 88, no. 2, 2010, DOI: 10.1007/s11263-009-0275-4.

[14] Davis, James W., and Mark A. Keck. "A two-stage template approach to person detection in thermal imagery." in proceedings of 2005 *Seventh IEEE Workshops on Applications of Computer Vision (WACV/MOTION'05)-Volume 1*, IEEE, 2005, pp. 364-369.

[15] Choi, Yukyung, Namil Kim, Soonmin Hwang, Kibaek Park, Jae Shin Yoon, Kyounghwan An, and In So Kweon. "KAIST multi-spectral day/night data set for autonomous and assisted driving." *IEEE Transactions on Intelligent Transportation Systems* 19, no. 3, 2018, pp: 934-948.

[16] FLIR Thermal Dataset. [Online] Available: 'https://www.flir.com/oem/adas/adas-dataset-form/', (Last accessed on 22nd February 2021).

[17] Xu, Fengliang, Xia Liu, and Kikuo Fujimura. "Pedestrian detection and tracking with night vision." *IEEE Transactions on Intelligent Transportation Systems* 6, no. 1, pp: 63-71, 2005, DOI: 10.1109/TITS.2004.838222, [Online].

[18] Olmeda, Daniel, Cristiano Premebida, Urbano Nunes, Jose Maria Armingol, and Arturo de la Escalera. "Pedestrian detection in far infrared images." *Integrated Computer-Aided Engineering* 20, no. 4, pp: 347-360, (2013), DOI: 10.3233/ICA-130441, [Online].

[19] S. L. Chang, F. T. Yang, W. P. Wu, Y. A. Cho, and S. W. Chen, "Nighttime pedestrian detection using thermal imaging based on HOG feature," 2011, doi: 10.1109/ICSSE.2011.5961992.

[20] N. Dalal and B. Triggs, "Histograms of oriented gradients for human detection," in *Proceedings - 2005 IEEE Computer Society Conference on Computer Vision and Pattern Recognition, CVPR 2005*, 2005, vol. I, doi: 10.1109/CVPR.2005.177.

[21] T. K. An and M. H. Kim, "A new Diverse AdaBoost classifier," in *Proceedings - International Conference on Artificial Intelligence and Computational Intelligence, AICI 2010*, 2010, vol. 1, doi: 10.1109/AICI.2010.82.

[22] Soundrapandiyan, Rajkumar, and PVSSR Chandra Mouli. "Adaptive pedestrian detection in infrared images using background subtraction and local thresholding." *Procedia Computer Science* 58, pp: 706-713, 2015.

[23] R. Lahmyed, M. El Ansari, and A. Ellahyani, "A new thermal infrared and visible spectrum images-based pedestrian detection system," Multimed. Tools Appl., vol. 78, no. 12, 2019, DOI: 10.1007/s11042-018-6974-5.

[24] A. Torabi, G. Massé, and G. A. Bilodeau, "An iterative integrated framework for thermal-visible image registration, sensor fusion, and people tracking for video surveillance applications," *Comput. Vis. Image Underst.*, vol. 116, no. 2, 2012, DOI: 10.1016/j.cviu.2011.10.006.

[25] J. Liu, S. Zhang, S. Wang, and D. N. Metaxas, "Multispectral deep neural networks for pedestrian detection," in *British Machine Vision Conference 2016, BMVC 2016*, 2016, vol. 2016-September, doi: 10.5244/c.30.73.

[26] J. Wagner, V. Fischer, M. Herman, and S. Behnke, "Multispectral pedestrian detection using deep fusion convolutional neural networks," 2016.

[27] M. Vandersteegen, K. Van Beeck, and T. Goedemé, "Real-Time Multispectral Pedestrian Detection with a Single-Pass Deep Neural Network," in *Lecture Notes in Computer Science (including subseries Lecture Notes in Artificial Intelligence and Lecture Notes in Bioinformatics)*, 2018, vol. 10882 LNCS, DOI: 10.1007/978-3-319-93000-8_47.

[28] D. Konig, M. Adam, C. Jarvers, G. Layher, H. Neumann, and M. Teutsch, "Fully Convolutional Region Proposal Networks for Multispectral Person Detection," in *IEEE Computer Society Conference on Computer Vision and Pattern Recognition Workshops*, 2017, vol. 2017-July, DOI: 10.1109/CVPRW.2017.36.

[29] D. Ghose, S. M. Desai, S. Bhattacharya, D. Chakraborty, M. Fiterau, and T. Rahman, "Pedestrian detection in thermal images using saliency maps," in *IEEE Computer Society Conference on Computer Vision and Pattern Recognition Workshops*,

[30] S. Ren, K. He, R. Girshick, and J. Sun, "Faster R-CNN: Towards Real-Time Object Detection with Region Proposal Networks," *IEEE Trans. Pattern Anal. Mach. Intell.*, vol. 39, no. 6, 2017, DOI: 10.1109/TPAMI.2016.2577031.

[31] C. Herrmann, T. Müller, D. Willersinn, and J. Beyerer, "Real-time person detection in low-resolution thermal infrared imagery with MSER and CNNs," in *Electro-Optical and Infrared Systems: Technology and Applications XIII*, 2016, vol. 9987, DOI: 10.1117/12.2240940.

[32] Q. Liu, Z. He, X. Li, and Y. Zheng, "PTB-TIR: A Thermal Infrared Pedestrian Tracking Benchmark," *IEEE Trans. Multimed.*, vol. 22, no. 3, 2020, DOI: 10.1109/TMM.2019.2932615.

[33] N. U. Huda, B. D. Hansen, R. Gade, and T. B. Moeslund, "The effect of a diverse dataset for transfer learning in thermal person detection," *Sensors (Switzerland)*, vol. 20, no. 7, 2020, DOI: 10.3390/s20071982.

[34] Z. Wu, N. Fuller, D. Theriault, and M. Betke, "A thermal infrared video benchmark for visual analysis," 2014, DOI: 10.1109/CVPRW.2014.39.

[35] F. Munir, S. Azam, M. A. Rafique, A. M. Sheri, and M. Jeon, "Thermal Object Detection using Domain Adaptation through Style Consistency," *arXiv*. 2020.

[36] R. McAllister *et al.*, "Concrete problems for autonomous vehicle safety: Advantages of Bayesian deep learning," in *IJCAI International Joint Conference on Artificial Intelligence*, 2017, vol. 0, DOI: 10.24963/ijcai.2017/661.

[37] Heliaus European Union Project, https://www.heliaus.eu/ (Last accessed on 20th February 2021).

[38] L. Bottou, "Large-scale machine learning with stochastic gradient descent," 2010, DOI: 10.1007/978-3-7908-2604-3_16.

[39] D. P. Kingma and J. L. Ba, "Adam: A method for stochastic optimization," 2015.

[40] Difference between Batch decent and Gradient Decent, [Online] Available: https://www.geeksforgeeks.org/difference-between-batch-gradient-descent-and-stochastic-gradient-descent/, (Last accessed on 20th February 2021).

[41] Pytorch deep learning framework, [Online] Available: https://pytorch.org/, (Last accessed on 14th October 2020).

[42] Nvidia Jetson Nano, [Online] Available: https://developer.nvidia.com/embedded/jetson-nano-developer-kit, (Last accessed on 14th January 2021).

[43] Nvidia TensorRT for developers, [Online] Available, https://developer.nvidia.com/tensorrt, (Last accessed on 14th February 2021).

[44] N. Wojke, A. Bewley, and D. Paulus, "Simple online and realtime tracking with a deep association metric," in *Proceedings - International Conference on Image Processing, ICIP*, 2018, vol. 2017-September, DOI: 10.1109/ICIP.2017.8296962.

[45] B. Besbes, A. Rogozan, A. M. Rus, A. Bensrhair, and A. Broggi, "Pedestrian detection in far-infrared daytime images using a hierarchical codebook of SURF," *Sensors (Switzerland)*, vol. 15, no. 4, 2015, DOI: 10.3390/s150408570.

[46] M. Teena and A. Manickavasagan, "Thermal infrared imaging," in *Imaging with Electromagnetic Spectrum: Applications in Food and Agriculture*, vol. 9783642548888, 2014.

[47] Lynred France, https://www.lynred.com/ (Last accessed on 20th March 2021).

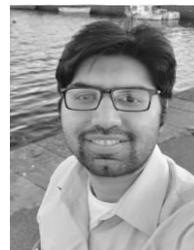

**Muhammad Ali Farooq** received his BE degree in electronic engineering from IQRA University in 2012 and his MS degree in electrical control engineering from the National University of Sciences and Technology (NUST) in 2017. He is currently pursuing the Ph.D. degree at the National University of Ireland Galway (NUIG). His research interests include machine vision, computer vision, video analytics, and sensor fusion. He has won the prestigious H2020 European Union (EU) scholarship and currently working as one of the consortium partners in the Heliaus (therrmal vision augmented awareness) project funded by EU.

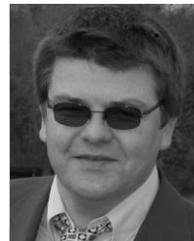

**Peter Corcoran** (Fellow, IEEE) holds a Personal Chair in Electronic Engineering at the College of Science and Engineering, National University of Ireland Galway (NUIG) . He was the Co-Founder in several start-up companies, notably FotoNation, now the Imaging Division of Xperi Corporation. He has more than 600 cited technical publications and patents, more than 120 peer-reviewed journal articles, 160 international conference papers, and a co-inventor on more than 300 granted U.S. patents. He is an IEEE Fellow recognized for his contributions to digital camera technologies, notably in-camera red-eye correction and facial detection. He is a member of the IEEE Consumer Technology Society for more than 25 years and the Founding Editor of IEEE Consumer Electronics Magazine.

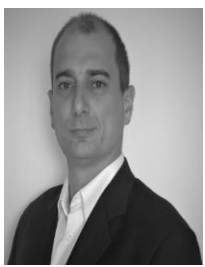

**Cosmin Rotariu** received the M.S. degree in medical informatics from National University of Ireland Galway (NUIG) in 2006. He is working as senior staff engineer in systems team at Xperi corporation Ireland. He is associated with industry for more than 12 years and holds vast experience in the areas of embedded systems designs, data filtering, compression, code optimizations, hardware design and firmware designs. He is involved in many cutting-edge technology projects with industry which includes driver and occupant monitoring systems (DOMS), wireless networks and low-level protocols designs for medical devices.